\documentclass{article}

\usepackage{microtype}
\usepackage{graphicx}
\usepackage{subfigure}
\usepackage{booktabs} % for professional tables
\usepackage[normalem]{ulem}

\usepackage{hyperref}

\usepackage[accepted]{icml2025}

\usepackage{amsmath}
\usepackage{amssymb}
\usepackage{mathtools}
\usepackage{amsthm}

\usepackage{multirow}

\usepackage[capitalize,noabbrev]{cleveref}
\newcommand{\std}[1]{\text{\scriptsize{(#1)}}}

\usepackage{array}
\newcolumntype{H}{>{\setbox0=\hbox\bgroup}c<{\egroup}@{}}

\theoremstyle{plain}

\theoremstyle{definition}

\theoremstyle{remark}

\usepackage[textsize=tiny]{todonotes}

\icmltitlerunning{QLESS: A Quantized Approach for Data Valuation and Selection in Large Language Model Fine-Tuning}

\begin{document}

\twocolumn[
\icmltitle{QLESS: A Quantized Approach for Data Valuation and Selection in Large Language Model Fine-Tuning}



\icmlsetsymbol{equal}{*}

\begin{icmlauthorlist}
\icmlauthor{Moses Ananta}{alpha}
\icmlauthor{Muhammad Farid Adilazuarda}{beta}
\icmlauthor{Zayd Muhammad Kawakibi Zuhri}{beta}
\icmlauthor{Ayu Purwarianti}{alpha}
\icmlauthor{Alham Fikri Aji}{beta}
\end{icmlauthorlist}

\icmlaffiliation{alpha}{School of Electrical Engineering and Informatics, Institut Teknologi Bandung, Bandung, Indonesia}
\icmlaffiliation{beta}{Mohamed bin Zayed University of Artificial Intelligence, Abu Dhabi, United Arab Emirates}
\icmlcorrespondingauthor{Moses Ananta}{23523016@std.stei.itb.ac.id}

\icmlkeywords{Machine Learning, ICML}

\vskip 0.3in
]



\printAffiliationsAndNotice{}  % leave blank if no need to mention equal contribution

\begin{abstract}
    Fine-tuning large language models (LLMs) is often constrained by the computational costs of processing massive datasets. We propose \textbf{QLESS} (Quantized Low-rank Gradient Similarity Search), which integrates gradient quantization with the LESS framework to enable memory-efficient data valuation and selection. QLESS employs a two-step compression process: first, it obtains low-dimensional gradient representations through LoRA-based random projection; then, it quantizes these gradients to low-bitwidth representations. Experiments on multiple LLM architectures (LLaMA, Mistral, Qwen) and benchmarks (MMLU, BBH, TyDiQA) show that QLESS achieves comparable data selection performance to LESS while reducing memory usage by up to 16x. Even 1-bit gradient quantization preserves data valuation quality. These findings underscore QLESS as a practical, scalable approach to identifying informative examples within strict memory constraints.\footnote{Code and data is available at \url{https://github.com/mosesananta/QLESS}}
\end{abstract}

\begin{figure}[t!]
    \centering
    \includegraphics[width=1\linewidth]{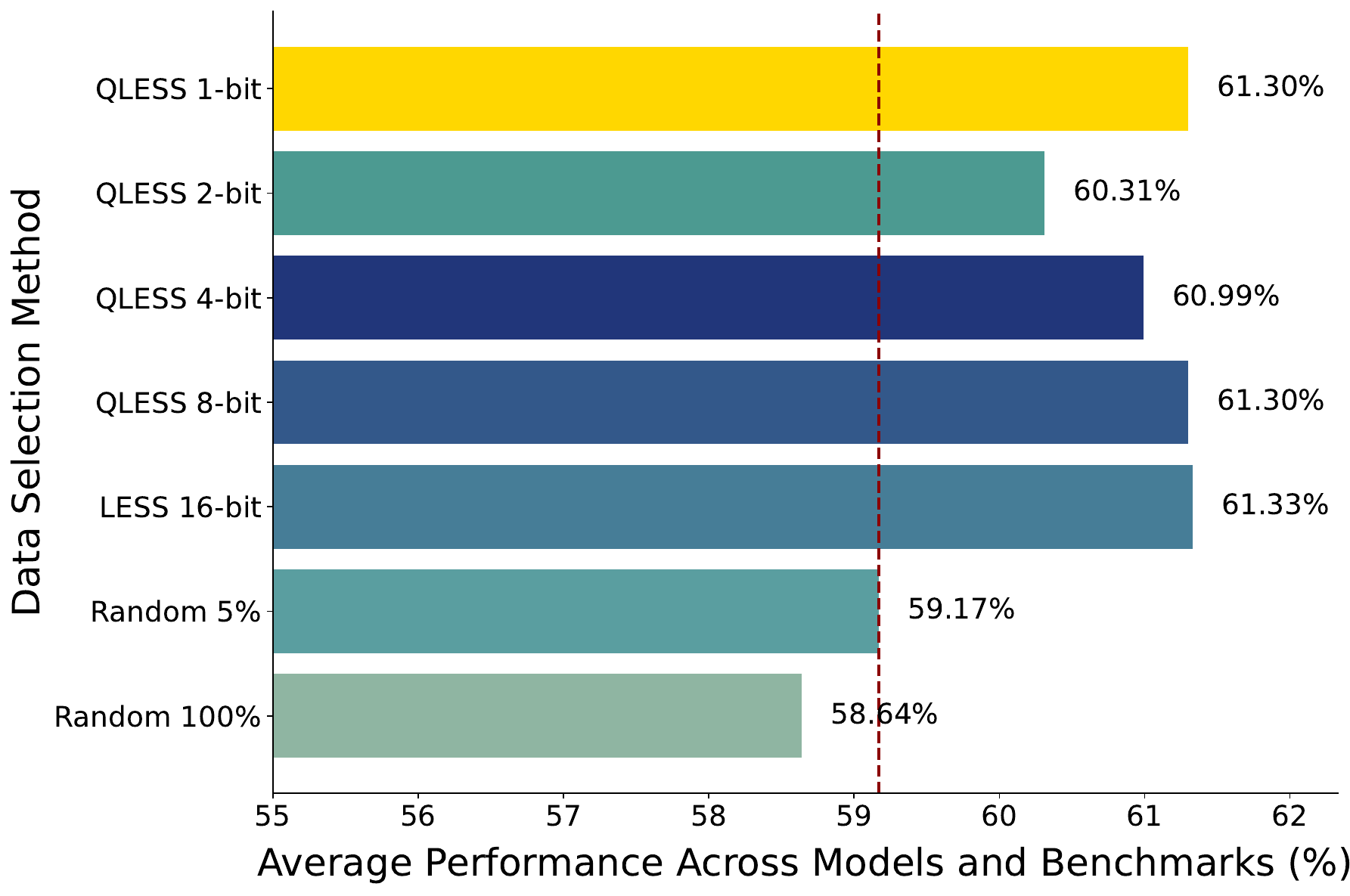}
    \caption{Data selection method comparison on average model performance across various models and benchmarks. The x-axis shows the average performance (\%), while the y-axis represents different data selection methods. QLESS with various quantization levels achieves comparable performance to LESS.}
    \label{fig:main-fig}
\end{figure}

\section{Introduction}

As Large Language Models (LLMs) continue to scale in size and complexity, their training and fine-tuning demand efficient data management strategies to optimize performance on specific tasks. A pivotal aspect of this optimization is data valuation—the process of identifying influential training data to improve learning outcomes. Gradient-based data valuation methods \cite{influence,tracin,trak}, which estimate the contribution of individual training samples to model performance, have proven effective in this regard. However, these approaches face significant scalability challenges: the storage and computation of high-precision gradient data grow disproportionately with the size of the model and dataset, making them impractical for modern LLMs \cite{DataInf,loGra,less}.

The LESS framework \cite{less} addresses some of these challenges by employing Low-Rank Adaptation (LoRA) for efficient gradient extraction and leveraging random projection to construct a compact gradient datastore. By normalizing and storing these reduced gradients, LESS enables computationally efficient similarity searches for data valuation. However, LESS retains a reliance on high-precision gradient storage, which imposes a considerable memory overhead, particularly for larger models and datasets. This bottleneck motivates a critical question: Can we significantly reduce the precision of stored gradients without sacrificing their utility for influence computation?

Inspired by gradient compression techniques in distributed and federated learning \cite{SignSGD,QSGD,TernGrad}, we propose \textbf{QLESS}—an extension of LESS that incorporates absmax-based quantization into the gradient datastore. QLESS replaces the high-precision gradient representation with quantized gradients, allowing us to drastically reduce memory requirements while maintaining the integrity of influence computations. Specifically, we employ a simple absmax-based uniform quantization scheme, which enables efficient storage of randomly projected gradients. The quantized gradients are then normalized during influence calculation, ensuring that the data valuation process remains robust despite reduced precision.

Our empirical findings demonstrate that QLESS achieves competitive fine-tuning performance even under extreme quantization levels. Notably, \textbf{1-bit quantized gradients}, despite their minimal precision, frequently perform on par with their 16-bit counterparts in data selection for instruction tuning. This remarkable result challenges conventional assumptions about the trade-offs between precision and efficacy in gradient-based data valuation and raises intriguing questions about the robustness of influence computations under extreme quantization.

In this paper, we make the following contributions:

\begin{itemize}
    \item \textbf{Quantized Gradient Datastore for Data Valuation:} We introduce a novel extension of LESS \cite{less} by integrating simple absmax-based quantization into the gradient datastore, achieving substantial memory savings without compromising the accuracy of influence computations.
    \item \textbf{Empirical Validation Across Models and Benchmarks:} Through extensive experiments on diverse training datasets (Flan v2 \cite{flanv2}, CoT \cite{COT}, Dolly \cite{Dolly}, and OpenAssistant \cite{OpenAssistant} ) and test datasets (MMLU \cite{mmlu}, BBH \cite{bbh}, and TyDiQA \cite{tydiqa}), we demonstrate the efficacy of QLESS across multiple quantization levels and LLM architectures, including LLaMA2 \cite{Llama2}, LLaMA3 \cite{Llama3}, Mistral \cite{Mistral7B}, and Qwen \cite{Qwen25} models.
    \item \textbf{Exploration of Quantization Trade-offs:} We analyze the impact of varying quantization levels (1 to 8-bit) on data valuation accuracy and downstream model performance, identifying cases where extreme quantization achieves unexpectedly robust results.
\end{itemize}

QLESS bridges the gap between efficiency and scalability in data valuation for LLMs, offering a practical and resource-efficient solution for instruction tuning.
%-------------------------------------------------------------
\section{Preliminaries}
\label{sec:preliminaries}

\subsection{Gradient-Based Data Valuation}
Data valuation methods aim to quantify each training sample’s contribution to a model’s performance. Among these, gradient-based approaches are particularly appealing for large-scale models, as they directly measure how training examples influence parameter updates. A key representative is TracIn~\cite{tracin}, which evaluates the impact of a training sample \(z\) on a validation sample \(z'\) by summing the dot products of their gradients across \(k\) checkpoints:
\[
\mathrm{Inf}_{\text{TracIn}}(z, z')
=
\sum_{i=1}^{k}
\eta_i\,
\langle
\nabla \ell(z; \theta_i),
\nabla \ell(z'; \theta_i)
\rangle,
\]
where \(\nabla \ell(z; \theta_i)\) and \(\nabla \ell(z'; \theta_i)\) are the gradients of the model’s loss at checkpoint \(i\), and \(\eta_i\) is the learning rate. This dot-product summation captures how \(z\) “pushes” the parameters in directions that aid (or impede) performance on \(z'\).  

However, storing gradients for multiple checkpoints becomes prohibitive in large language models (LLMs). To address this, approaches like \textbf{LESS}~\cite{less} apply compression strategies that retain crucial gradient alignments while reducing the memory burden.

\begin{figure*}[ht!]
\vskip 0.1in
\begin{center}
\centerline{\includegraphics[width=\textwidth]{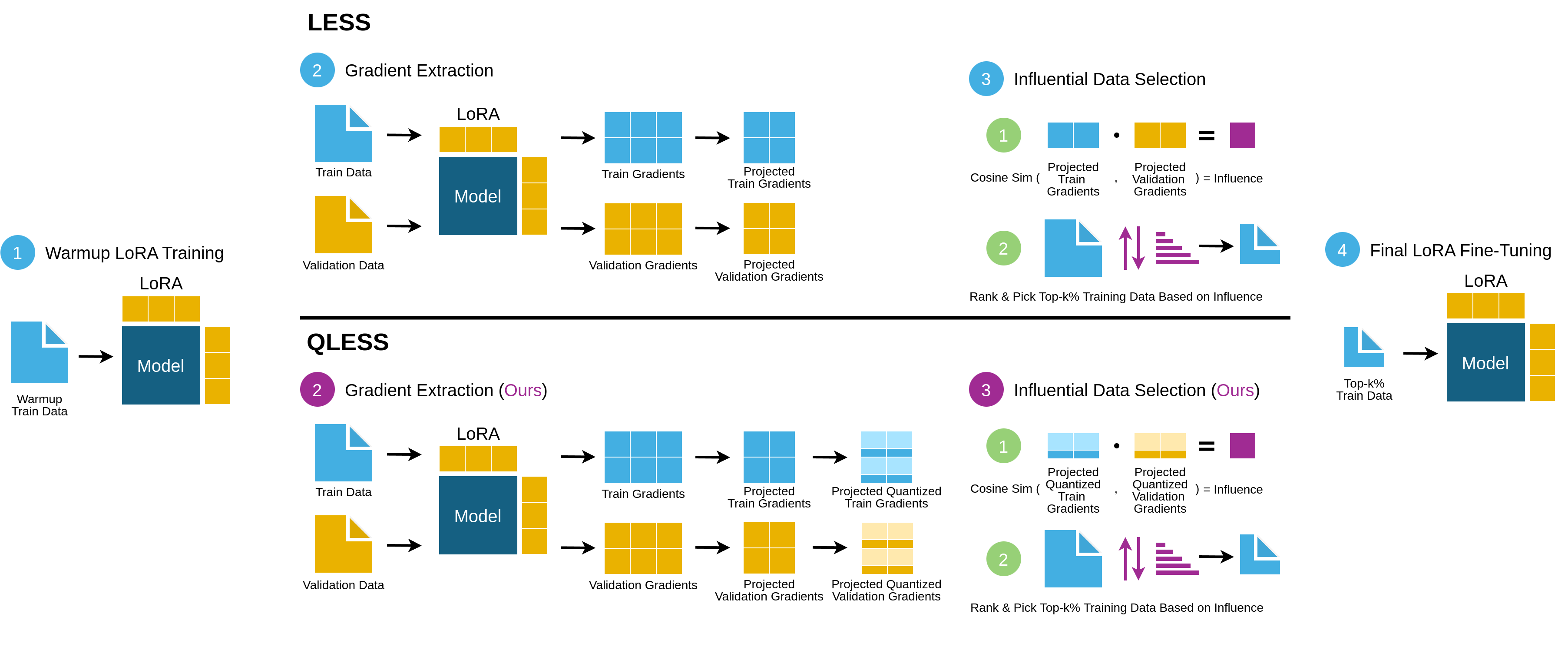}}
\caption{Overview of the data valuation and selection process. LESS and QLESS differ from each other in step 2 and 3.}
\label{fig:qless-flowchart}
\end{center}
\vskip -0.1in
\end{figure*}

\subsection{LESS: Low-Rank Gradient Similarity Search}
\label{sec:less}
\textbf{LESS} (Low-Rank Gradient Similarity Search)~\cite{less} extends gradient-based data valuation by leveraging the parameter efficiency of LoRA~\cite{LoRA}. Instead of computing full-model gradients, LESS focuses on the \textit{LoRA} parameters, extracting Adam training gradient 
\(\Gamma(z; \theta)\)
for each training sample \(z\). For validation samples \(z'\), it uses an SGD validation gradient 
\(\nabla \ell(z'; \theta)\)
to measure how training samples influence validation performance.

To handle the high dimensionality of these gradients, LESS applies a randomly initialized linear mapping 
\(R: \mathbb{R}^d \to \mathbb{R}^k\)
where \(k \ll d\). By the Johnson--Lindenstrauss lemma~\cite{JohnsonLindenstrauss}, this projection approximately preserves pairwise distances or inner products, enabling efficient similarity searches. Concretely, for each training sample \(z\) at checkpoint \(i\):
\begin{equation}
g_{z,i} = R\bigl(\Gamma(z; \theta_i)\bigr) \;\in\;\mathbb{R}^k.
\end{equation}

In instruction tuning, each sample \(z\) often spans multiple tokens, so its gradient is an average of token-level gradients. As reported in~\cite{less}, this can disproportionately favor shorter sequences by yielding larger gradient norms. To counteract this bias, LESS normalizes each projected gradient:
\begin{equation}
\hat{g}_{z,i} = \frac{g_{z,i}}{\|g_{z,i}\|},
\end{equation}
and stores \(\hat{g}_{z,i}\) in a gradient datastore. During inference, the influence of \(z\) on a validation sample \(z'\) is then computed by summing the learning-rate-weighted cosine similarities across \(N\) checkpoints:
\begin{align}
\mathrm{Inf}_{\text{LESS}}(z, z') 
&= \sum_{i=1}^{N} \eta_i 
\;\frac{ 
   \langle 
      \nabla \ell(z'; \theta_i), 
      \Gamma(z; \theta_i)
   \rangle
}{
   \|\nabla \ell(z'; \theta_i)\|\,
   \|\Gamma(z; \theta_i)\|
}
\nonumber \\
&= \sum_{i=1}^{N} \eta_i \,\bigl\langle \hat{g}_{z',i}, \hat{g}_{z,i} \bigr\rangle.
\end{align}

Although LESS significantly reduces dimensionality and computation, storing \(\hat{g}_{z,i}\) in floating-point (or even half-precision) remains costly for large datasets, motivating further compression via quantization.

\subsection{Quantization in Distributed and Federated Learning}
\label{sec:quant_background}
Quantization---mapping high-precision numerical data to lower bit-width representations---has been widely adopted in distributed and federated learning to mitigate communication overhead~\cite{SignSGD,QSGD,TernGrad}. The core idea is to transform a floating-point vector $\mathbf{x} \in \mathbb{R}^k$ into a more compact integer or fixed-point format, frequently reducing memory by factors of four, eight, or more.

A straightforward technique is absmax-based quantization, which scales each vector by its largest absolute value to fit into a specified bit budget $b$. Suppose
\begin{equation}
S 
= 
\max_{1 \le m \le k}
\bigl|
x_m
\bigr|,
\end{equation}
and let $\alpha = 2^{b-1} - 1$. Each component $x_m$ is mapped to an integer $q_m$ via
\begin{equation}
q_m
=
\mathrm{round}\!\Bigl(
\alpha
\,\frac{x_m}{S}
\Bigr).
\end{equation}
This compression technique reduces the memory footprint dramatically, facilitating the storage or transfer of gradient-related data in large-scale systems.

\subsection{Quantized Random Projection (QRP)}
\label{sec:quant_rp}

Quantized random projection (QRP) combines the dimensionality reduction benefits of random projection with discrete quantization to produce compact vector representations while preserving key geometric relationships. The theoretical foundations of QRP lie in the Johnson--Lindenstrauss lemma~\cite{JohnsonLindenstrauss} and its quantized variants~\cite{QuantizedJL,QRPCosine,QRPGuarantee}, which provide guarantees on the approximate preservation of pairwise distances or inner products under random projection and quantization. 

In the conventional random projection pipeline, we map high-dimensional vectors 
\(\{\mathbf{x}_i\}_{i=1}^n \subset \mathbb{R}^d\)
to a lower-dimensional space 
\(\mathbb{R}^k\)
via a randomly sampled matrix 
\(\mathbf{R} \in \mathbb{R}^{k \times d}\)
drawn from a suitable distribution (e.g., Gaussian or Rademacher). 

Once the vectors \(\mathbf{y}_i = \mathbf{R}\mathbf{x}_i\) are projected into \(\mathbb{R}^k\), a quantization function
\[
Q: \mathbb{R}^k \to \mathcal{C}^k
\]
further compresses them into a set of finite levels \(\mathcal{C}\). For instance, binary QRP restricts each dimension to \(\{-1, +1\}\)~\cite{Simhash,VerySparseRP}, whereas uniform scalar quantization maps each dimension to uniformly spaced bins~\cite{QRPImage,RPCoding} . The resulting codes \(\mathbf{q}_i = Q(\mathbf{y}_i)\) can be stored efficiently in low-precision formats, leading to substantial memory and computational savings. %Prior work has shown QRP to be effective in applications such as similarity search~\cite{QRPsearch}, clustering~\cite{QRPclustering}, and compressed learning~\cite{QRPlearning}.

In gradient-based data valuation for large language models, the principal challenge lies in storing and manipulating massive gradient datasets. Although methods like LESS~\cite{less} reduce dimensionality via random projection, they typically retain a floating-point representation, which can still be expensive at scale. QRP provides an additional layer of compression on top of random projection, ensuring that large sets of projected gradients remain compact.

%-------------------------------------------------------------
\section{QLESS: Integrating LESS with Quantization}
\label{sec:methodology}

Building upon the gradient-based data valuation strategies introduced in the previous section, we now present \textbf{QLESS}, a method that combines the memory-efficient random projection and normalization of LESS with an additional quantization step. The key insight is that even after reducing gradient dimensionality, storing the resulting vectors in floating-point format can be expensive for large datasets. By quantizing each projected gradient, QLESS substantially reduces memory consumption while preserving enough directional information to maintain the effectiveness of influence-based data selection. The full process is visualized in Figure~\ref{fig:qless-flowchart}.

\subsection{Quantizing Randomly Projected Gradients}
\label{sec:quantizing_rp_gradients}

Building on the random projection and normalization strategy in LESS~\cite{less}, QLESS further leverages Quantized Random Projection (QRP; see Section~\ref{sec:quant_rp}) to drastically reduce the memory footprint of stored gradients. Specifically, for each training sample \(z\) at checkpoint \(i\), QLESS computes a LoRA-based gradient \(\Gamma(z; \theta_i)\) and applies the same random projection
\[
g_{z,i} \;=\; R\bigl(\Gamma(z; \theta_i)\bigr) \;\in\; \mathbb{R}^k
\]
as in LESS. However, rather than storing \(g_{z,i}\) in floating-point format, QLESS immediately quantizes it using a simple but effective absmax-based scheme, where vectors are compressed after random projection to retain essential geometric information.

Concretely, let
\[
S_{z,i} = \max_{1 \le m \le k} \bigl|g_{z,i,m}\bigr|,
\quad
\alpha = 2^{b-1} - 1,
\]
for a chosen bit budget \(b \in \{1,2,4,8\}\). Each component \(g_{z,i,m}\) is mapped to an integer \(q_{z,i,m}\) via
\[
q_{z,i,m} \;=\; \mathrm{round}\!\Bigl(
  \alpha \,\frac{g_{z,i,m}}{S_{z,i}}
\Bigr).
\]
QLESS stores the integer vector \(q_{z,i}\) together with the single scale factor \(S_{z,i}\), reducing memory costs from \(k\) floating-point values to \(k\) \(b\)-bit integers plus one float per gradient vector.

\subsection{Influence Computation with Quantized-Normalized Gradients}
During data valuation, QLESS normalizes the quantized representation of \(g_{z,i}\) directly. Each quantized component \(q_{z,i,m}\) is normalized as:
\begin{equation}
\widehat{q}_{z,i,m} = \frac{q_{z,i,m}}{\|\mathbf{q}_{z,i}\|},
\end{equation}
where \(\|\mathbf{q}_{z,i}\|\) is the norm of the quantized vector.

Similarly, validation gradients \(g_{z',i}\) are quantized and normalized, yielding \(\widehat{q}_{z',i}\). QLESS computes the influence of a training sample \(z\) on a validation sample \(z'\) by summing the learning-rate-weighted cosine similarities between the quantized-normalized gradients across \(N\) checkpoints:
\begin{equation}
\mathrm{Inf}_{\text{QLESS}}(z, z') = \sum_{i=1}^{N} \eta_i \;\bigl\langle \widehat{q}_{z',i}, \widehat{q}_{z,i} \bigr\rangle.
\end{equation}

By utilizing QRP (\textsection\ref{sec:quantizing_rp_gradients}), QLESS retains the essence of gradient-based data valuation while substantially cutting storage demands. As we will show in our experiments, this approach enables efficient data selection across larger datasets and a broader range of LLM architectures and sizes than would be practical with higher-precision storage.

%-------------------------------------------------------------
\section{Experiments}
\label{sec:experiments}

This section describes our experimental setup for evaluating QLESS, focusing on datasets, models, baselines, and the overall pipeline. By adhering to the same framework as LESS, we ensure fair comparisons while isolating the effects of gradient quantization introduced in QLESS.

\begin{table*}[ht!]
\caption{Performance of Qwen 2.5-7B and Llama 3.1-8B models with different data selection methods and gradient storage space. Metrics include TyDiQA, MMLU, BBH, and their average in percent (\%). \textbf{Bold} and \underline{underline} denote the best and second best performing setup across selection methods of the same model.}
\label{table:results}
\centering
\small
\vspace{3mm}
\begin{tabular}{lcccccccc}
\toprule
\textbf{Model} & \textbf{Data Selection} & \textbf{Storage} & \textbf{TyDiQA} & \textbf{MMLU} & \textbf{BBH} & \textbf{Avg} \\
\midrule
\multirow{7}{*}{Qwen 2.5 7B} 
  & random 100\% & - & 69.24 & \textbf{74.20} & 66.76 & 70.06 \\
  & random 5\%  & - & 70.35 \std{2.2} & \underline{74.10} \std{0} & 66.60 \std{0.9} & 70.35 \std{0.5} \\
  & LESS 16-bit & 16.54 GB & \textbf{72.36} \std{1.2} & 74.10 \std{0.1} & 67.44 \std{0.9} & \textbf{71.30} \std{0.4} \\
  & QLESS 8-bit & 8.27 GB & 70.96 \std{2.6} & 73.97 \std{0.2} & \textbf{67.99} \std{0.7} & 70.97 \std{1.0} \\
  & QLESS 4-bit & 4.14 GB & 70.89 \std{1.9} & 74.03 \std{0.1} & \underline{67.69} \std{1.0} & 70.87 \std{0.8} \\
  & QLESS 2-bit & 2.07 GB & \underline{72.13} \std{0.4} & 74.07 \std{0.1} & 67.25 \std{0.6} & \underline{71.15} \std{0.1} \\
  & QLESS 1-bit & 1.03 GB & 70.97 \std{0.4} & 74.00 \std{0.1} & 67.19 \std{1.6} & 70.72 \std{0.7} \\
\midrule
\multirow{7}{*}{Llama 3.1 8B} 
  & random 100\% & - & 67.91 & 64.50 & \textbf{65.37} & 65.93 \\
  & random 5\%  & - & 65.65 \std{3.4} & \textbf{65.47} \std{0.1} & 63.02 \std{1.0} & 64.71 \std{1.4} \\
  & LESS 16-bit & 16.54 GB & \textbf{68.89} \std{2.3} & 64.93 \std{0.5} & 64.85 \std{4.0} & \underline{66.22} \std{2.2} \\
  & QLESS 8-bit & 8.27 GB & \underline{68.83} \std{2.4} & \underline{65.03} \std{0.4} & 64.75 \std{4.0} & 66.20 \std{2.2} \\
  & QLESS 4-bit & 4.14 GB & 68.77 \std{3.3} & 64.87 \std{0.5} & \underline{65.31} \std{4.1} & \textbf{66.31} \std{2.6} \\
  & QLESS 2-bit & 2.07 GB & 68.58 \std{2.2} & 64.90 \std{0.4} & 63.95 \std{2.7} & 65.81 \std{1.7} \\
  & QLESS 1-bit & 1.03 GB & 67.84 \std{3.2} & 64.97 \std{0.6} & 64.97 \std{3.6} & 65.93 \std{2.4} \\
\bottomrule
\end{tabular}
\end{table*}

\subsection{Experimental Setup}
\label{subsec:setup}

\paragraph{Datasets.} We adopt the same datasets used in LESS to ensure consistency.  The training data comprises four major instruction-tuning datasets: Flan v2 (100K examples) \cite{flanv2}, Chain-of-Thought (100k examples) \cite{COT}, Dolly (15k examples) \cite{Dolly}, and OpenAssistant (55K examples) \cite{OpenAssistant}, totaling approximately 270K training instances. These datasets span diverse instruction types, from single-turn queries to complex reasoning tasks and multi-turn conversations. For validation and evaluation, we use the same benchmarks as LESS: MMLU \cite{mmlu}, which tests knowledge across 57 academic subjects, BBH \cite{bbh} for assessing complex reasoning capabilities, and TyDiQA \cite{tydiqa} for evaluating cross-lingual understanding across 11 typologically diverse languages. Following LESS, we construct our validation set $\mathcal{D}_{\text{val}}$ using few-shot samples from these benchmarks to guide the data selection process.

\paragraph{Models.}
We evaluate QLESS on five language models spanning different architectures and parameter scales: LLaMA2-7B \cite{Llama2}, Mistral-7B \cite{Mistral7B}, Qwen2.5-7B \cite{Qwen25}, LLaMA3.1-8B, and LLaMA3.2-3B \cite{Llama3}. This diverse selection enables us to assess the generalizability of our approach across different model families and sizes.

\paragraph{Baselines.}
We compare QLESS against the following baselines:
\begin{enumerate}
    \item \textbf{Random Selection:} Randomly sample 5\% of the training dataset for instruction tuning. This serves as a lower bound for performance.
    \item \textbf{LESS (Original):} The original LESS framework serves as the main baseline. It involves:
    \begin{enumerate}
        \item \textbf{Warmup Training:} Perform warmup training on a randomly selected 5\% subset of the training data $\mathcal{D}_{\text{warmup}}$ for $N=4$ epochs.
        \item \textbf{Gradient Feature Extraction:} Extract 8192-dimensional gradient features for all training data $\mathcal{D}$ using LoRA-based adaptation with Adam.
        \item \textbf{Data Scoring and Selection:} Compute influence scores by measuring gradient alignment with validation gradients. Select the top 5\% of training data $\mathcal{D}_{\text{train}}$ based on these scores.
    \end{enumerate}
\end{enumerate}

\paragraph{Our Method.}
QLESS follows the same experimental pipeline as LESS to ensure a direct and fair comparison. The key distinction is in the gradient datastore construction step:
\begin{itemize}
    \item Projected gradients are quantized using absmax-based scaling, reducing storage from 16-bit floating-point precision to $b$-bit integers, where $b \in \{8, 4, 2, 1\}$.
    \item During influence computation, the quantized gradients are reconstructed, normalized, and compared with validation gradients using cosine similarity.
    \item The top 5\% of training samples $\mathcal{D}_{\text{train}}$ are selected based on cumulative influence scores.
\end{itemize}

\paragraph{Implementation Details.}
All experiments are conducted on 4 NVIDIA A100 GPUs (40GB). Hyperparameters, optimizer settings, learning rates, and batch sizes are identical to LESS and are detailed in Appendix~\ref{app:experimental_details} for reproducibility. 

\begin{table}[h!]
\caption{Performance of Qwen 2.5 7B under different model and gradient quantization settings across TyDiQA, MMLU, and BBH. The final column presents the average performance across these tasks. \textbf{Bold} and \underline{underline} denote the best and second best performing setup across model quantization and gradient quantization.}
\label{tab:qwen-2-5-7b-quant}
\vspace{3mm}
\centering
\small
\begin{tabular}{@{ }c@{ }cHcccc@{ }} 
\toprule
\textbf{Model Q} & \textbf{Grad Q} & \textbf{Memory} & \textbf{TyDiQA} & \textbf{MMLU} & \textbf{BBH} & \textbf{Avg} \\ 
\textbf{(Mem.)} \\ 
\midrule
– & rand 100\% & – & 69.24 & \underline{74.20} & 66.76 & 70.06 \\
– & rand 5\% & – & 72.45 & 74.10 & 66.02 & 70.86 \\ 
\cmidrule(lr){1-7}
\multirow{5}{*}{\shortstack{16-bit \\ (35.42 GB)}} & 16-bit & \multirow{5}{*}{35.42 GB} & 72.93 & 74.20 & 67.96 & 71.70 \\
 & 8-bit &  & \underline{73.65} & 73.80 & 67.87 & 71.77 \\
 & 4-bit &  & 73.05 & 74.00 & \textbf{68.43} & \textbf{71.83} \\
 & 2-bit &  & 72.55 & 74.10 & 66.76 & 71.14 \\
 & 1-bit &  & 71.11 & 74.00 & 68.52 & 71.21 \\ 
\cmidrule(lr){1-7}
\multirow{5}{*}{\shortstack{8-bit \\ (34.10 GB)}} & 16-bit & \multirow{5}{*}{34.10 GB} & 72.26 & 74.10 & 68.52 & 71.62 \\
 & 8-bit &  & 73.08 & 73.90 & 67.31 & 71.43 \\
 & 4-bit &  & 72.60 & 74.00 & 67.69 & 71.43 \\
 & 2-bit &  & 71.87 & 74.10 & 65.65 & 70.54 \\
 & 1-bit &  & 70.93 & 74.00 & 67.50 & 70.81 \\ 
\cmidrule(lr){1-7}
\multirow{5}{*}{\shortstack{4-bit \\ (26.09 GB)}} & 16-bit & \multirow{5}{*}{26.09 GB} & 73.12 & 74.10 & 68.24 & 71.82 \\
 & 8-bit &  & \textbf{73.81} & \textbf{74.30} & 67.78 & \uline{71.96} \\
 & 4-bit &  & 72.13 & \underline{74.20} & \uline{68.61} & 71.65 \\
 & 2-bit &  & 69.87 & 74.10 & 67.41 & 70.46 \\
 & 1-bit &  & 72.60 & 74.00 & 67.22 & 71.27 \\
\bottomrule
\end{tabular}
\end{table}

\subsection{Main Results}
\label{sec:results}

The results of our experiments, as summarized in Table~\ref{table:results} and Table~\ref{table:additional-model-table}, highlight the efficacy of QLESS in balancing storage efficiency and downstream model performance across various bit-width configurations. Our analysis focuses on the trade-offs between performance and storage, the generalizability of QLESS across models, and the surprising robustness of extreme quantization.

\paragraph{Overall Performance.} Our experiments demonstrate that QLESS achieves competitive performance across all evaluated models and benchmarks while significantly reducing storage requirements compared to LESS. On most benchmarks, QLESS results in comparable or better performance than LESS and the random baselines. Most notably, even with 1-bit quantization, Llama 3.1 8B for example achieves an average score of 65.93, exceeding the random 5\% baseline of 64.71, demonstrating the robustness of our approach under extreme quantization.

\paragraph{Performance vs. Storage Trade-Off.}
QLESS demonstrates a notable reduction in memory requirements while retaining competitive performance relative to LESS, particularly at higher bit-width configurations. For instance, 8-bit QLESS achieves results comparable to LESS on most benchmarks, despite cutting the gradient datastore size from 16.54\,GB to around 8.27\,GB. As the bit-width is lowered to 4-bit and 2-bit, performance declines moderately but still yields substantial storage savings (around 4.14\,GB and 2.07\,GB, respectively). Even at 1-bit precision, QLESS remains only a few points behind 16-bit performance while requiring just 1.03\,GB of storage, indicating that coarse directional information can be enough for effective data valuation.

\paragraph{Robustness Across Benchmarks.}
Across TyDiQA, MMLU, and BBH, QLESS at 8-bit precision frequently matches LESS, and 4-bit or 2-bit configurations remain within a tolerable performance gap. Although BBH can be more sensitive to quantization, 1-bit QLESS still performs competitively, confirming its resilience. These results underscore that even extreme compression maintains sufficient gradient information for robust data selection.

\paragraph{Insights into 1-Bit Quantization.}
A particularly striking outcome is the effectiveness of 1-bit QLESS, which often attains performance near that of higher-bit variants. This finding challenges the assumption that low-precision representations inevitably undermine gradient-based methods. Our evidence suggests that a uniform quantization scale can preserve enough directional gradient information for credible influence estimation, even under the most aggressive compression setting.

\begin{table}[ht!]
\caption{Performance of Llama 2 7B under different quantization methods across TyDiQA, MMLU, and BBH. The final column presents the average performance across these tasks. \textbf{Bold} and \underline{underline} denote the best and second best performing setup across quantization methods.}
\label{tab:absmean-v-absmax}
\vspace{3mm}
\centering
\small
\begin{tabular}{@{ }c@{ }ccccc@{ }} 
\toprule
\textbf{Q Scheme} & \textbf{Grad Q} & \textbf{TyDiQA} & \textbf{MMLU} & \textbf{BBH} & \textbf{Avg} \\ 
\cmidrule(lr){1-6}
– & rand 100\% & 49.89 & 47.30 & \textbf{40.93} & 46.04 \\
– & rand 5\% & 48.13 & 45.00 & 38.70 & 43.94 \\ 
\cmidrule(lr){1-6}
– & 16-bit & 54.97 & 47.70 & \uline{40.74} & 47.80 \\ 
\cmidrule(lr){1-6}
\multirow{3}{*}{Absmax} & 8-bit & \textbf{57.81} & 47.40 & 40.56 & 48.59 \\
 & 4-bit & 55.75 & 47.50 & 40.28 & 47.84 \\
 & 2-bit & 54.48 & 46.80 & 40.46 & 47.25 \\ 
\cmidrule(lr){1-6}
\multirow{3}{*}{Absmean} & 8-bit & 56.54 & \uline{47.90} & 40.19 & 48.21 \\
 & 4-bit & \uline{57.25} & 47.80 & 40.56 & 48.53 \\
 & 2-bit & 56.11 & \textbf{48.00} & 40.09 & 48.07 \\ 
\cmidrule(lr){1-6}
Sign & 1-bit & 56.93 & \textbf{48.00} & 40.46 & 48.46 \\
\bottomrule
\end{tabular}
\end{table}

\section{Ablation Studies and Analysis}
\paragraph{Effect of QLoRA on Gradient Extraction.}
Motivated by the desire to further reduce memory usage during gradient extraction, we replace the standard LoRA with QLoRA \cite{QLoRa} in our QLESS pipeline. Specifically, for model quantization, we employ the LLM int8 approach (8-bit) \cite{LLMint8} and bitsandbytes NF4 (4-bit) \cite{QLoRa}, enabling more efficient parameter storage while still preserving key model capabilities. Our primary question is whether QLoRA—when combined with gradient quantization—meaningfully degrades the gradient fidelity required for accurate influence estimation.

As shown in Table~\ref{tab:qwen-2-5-7b-quant} and Table~\ref{tab:llama-2-7b-quant}, the memory footprint drops from around 32–35 GB down to 22–26 GB for both LLaMA2-7B and Qwen2.5-7B, yet the average performance on TyDiQA, MMLU, and BBH remains close to that of higher-precision baselines. Although there is a minor drop in scores when combining QLoRA with 1-bit or 2-bit gradient quantization, the overall performance remains well above random-selection levels. These results suggest that using QLoRA in tandem with QLESS provides a viable solution for resource-constrained environments, where aggressive quantization of both model and gradients can significantly cut memory costs while preserving sufficient directional information for robust data selection.

\begin{figure}[ht]
\vskip 0.1in
\begin{center}
\centerline{\includegraphics[width=\columnwidth]{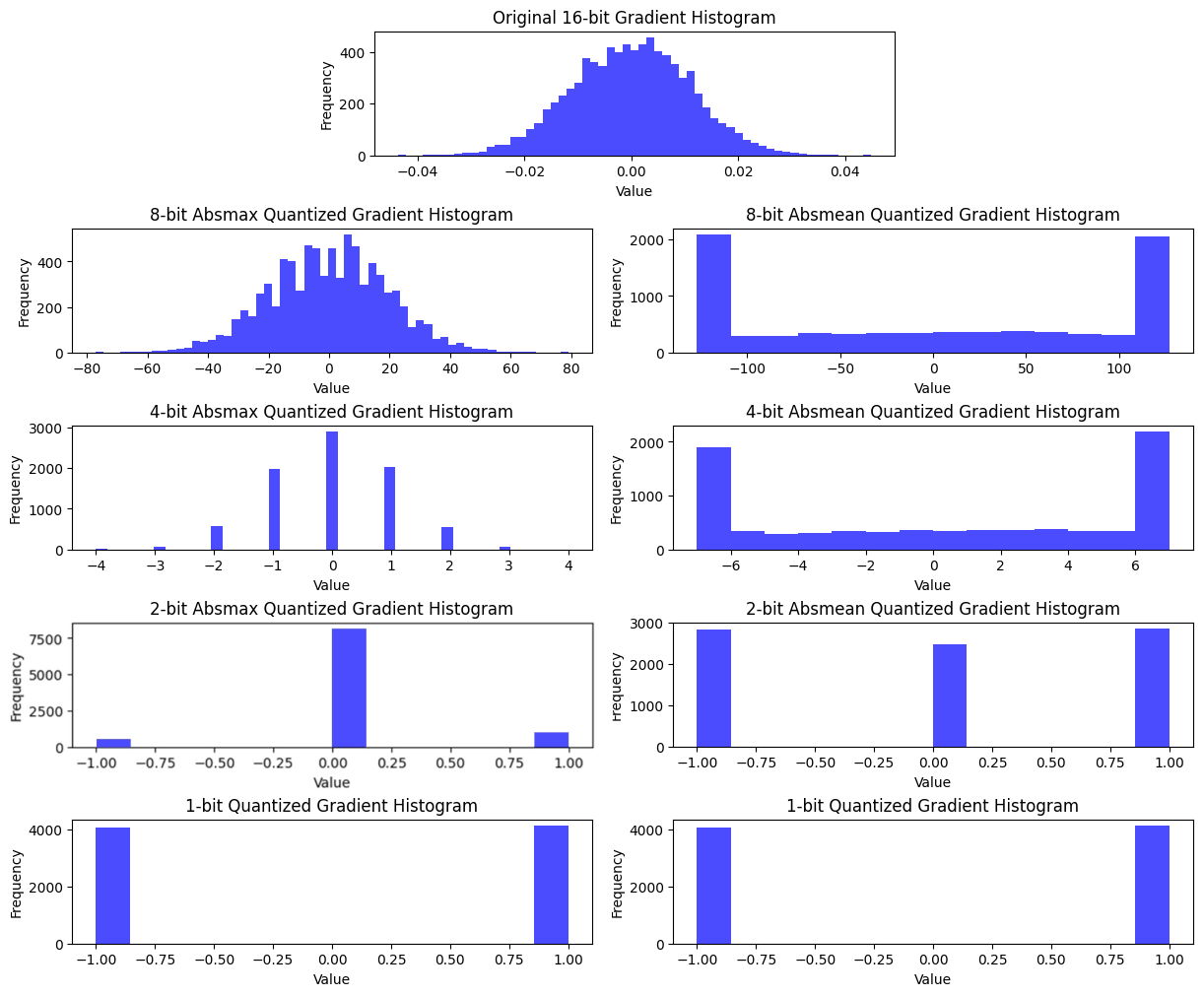}}
\caption{Absmax- vs absmean-based quantization comparison based on the distribution of values in the gradient tensors.}
\label{fig:absmean-v-absmax}
\end{center}
\vskip -0.1in
\end{figure}

\paragraph{Analyzing Sparsity Effects of Absmax Quantization}
While absmax-based quantization effectively compresses gradients, it can induce significant sparsity as bit-width decreases. In particular, most gradient values collapse into the zero bin for 2-bit and 4-bit settings, leading to a heavily sparse representation that degrades dot-product operations crucial for influence computation. This effect arises because the Gaussian-like distribution of gradients under absmax scaling often places a large fraction of values near zero, exacerbating the loss of directional information.

To counteract this, we explore absmean-based quantization, which shifts the quantization scale to emphasize the mean absolute value rather than the maximum. By pushing values away from zero and closer to the outer bins, absmean-based quantization yields a denser distribution of nonzero entries, thereby preserving more effective gradient directions for similarity calculations. Interestingly, 1-bit quantization does not exhibit the extreme sparsity problem because its representation inherently omits a zero bin. Figure~\ref{fig:absmean-v-absmax} provides visualizations showing how absmax-based quantization leads to severe zero bin occupancy in low-bit regimes and how absmean-based quantization alleviates this issue at the cost of reducing the fidelity of gradient information, especially in 8-bit and 4-bit precision. 

Table~\ref{tab:absmean-v-absmax} compares absmean and absmax quantization for LLaMA2-7B under various bit-widths. At higher precisions (e.g., 16-bit), absmax slightly outperforms absmean on TyDiQA and BBH, resulting in a higher overall average (48.59 vs.\ 47.80). However, as bit-width decreases to 4-bit and 2-bit, absmean consistently yields better or on-par results across all benchmarks, notably improving TyDiQA scores and maintaining competitive performance on MMLU and BBH. This reversal suggests that absmean’s tendency to push values away from zero is more robust when the quantization granularity is coarse, mitigating the zero-bin effect that degrades similarity computation. 

\begin{figure}[ht]
\vskip 0.1in
\begin{center}
\centerline{\includegraphics[width=\columnwidth]{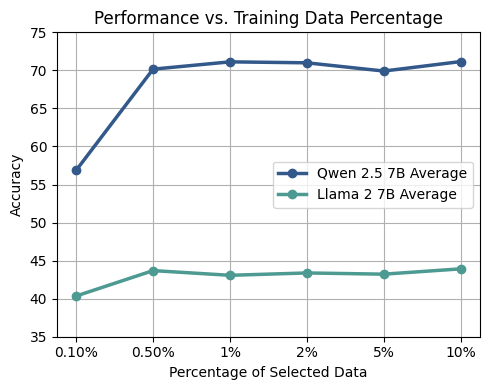}}
\caption{The average benchmark results of Llama 2 7B and Qwen 2.5 7B when fine-tuned on different percentages of QLESS selected data with 1-bit gradient store.}
\label{fig:percentages}
\end{center}
\vskip -0.1in
\end{figure}

\paragraph{Empirical results on selected data percentage.}
Our main experiments follow LESS~\cite{less} in selecting only 5\% of data for fair comparisons, although there has yet been empirical results on whether or not this is the optimal percentage. We experiment on selecting 0.1\%, 0.5\%, 1\%, 2\%, and 10\% in addition to 5\% of data, selected using QLESS specifically at 16-bit model precision and 1-bit gradient store quantization using Llama 2 7B and Qwen 2.5 7B on the 3 benchmarks.

The results can be seen in Figure~\ref{fig:percentages}. Interestingly, we find that the performance of the models are already obtainable even at 0.5\%. Selecting only 0.1\% may not be enough, but performance on these benchmarks plateaus starting from 0.5\% up to 5\% and only goes up slightly at 10\%. These results suggest that even with a tighter budget, it is still possible to gain performance with QLESS to some degree.

\section{Related Work}
\label{sec:related_work}

\paragraph{Gradient-based data valuation and selection.}
Data valuation techniques have evolved from computationally intensive leave-one-out approaches to gradient-based strategies~\cite{influence}. TracIn~\cite{tracin} measures a training sample’s impact on validation by aligning gradients at multiple checkpoints, offering a tractable alternative to second-order influence functions. Subsequent work refines gradient-based valuation to handle larger models and datasets, incorporating methods such as random projection~\cite{trak,less} and LoRA-based efficient gradient extraction~\cite{loGra,DataInf,less}. Our QLESS framework furthers this line of research by adding quantization to the gradient datastore, reducing memory overhead without compromising selection effectiveness.

\paragraph{Efficient fine-tuning and adaptation of LLMs.}
Parameter-efficient fine-tuning techniques reduce the overhead of adapting large models by training only a small fraction of parameters. LoRA~\cite{LoRA}, QLoRA~\cite{QLoRa}, and various adapter-based methods~\cite{AdaLoRA,LoHa,LoKr,OFT,BOFT} significantly lower GPU memory usage during specialization. QLESS leverages these developments by integrating parameter-efficient gradient extraction with quantization-based compression, allowing for more resource-friendly data selection in instruction tuning scenarios.

\paragraph{Gradient and model compression.}
Minimizing communication overhead in distributed and federated learning has sparked a range of gradient compression methods~\cite{Grad8bit,SignSGD,TernGrad,QSGD,SparseGD1,SparseGD2,SparseGD3}, which reduce precision or introduce sparsity while preserving convergence quality. These strategies are critical for large-scale model training where exchanging high-precision gradients is infeasible. Inspired by these insights, our approach applies quantization to the gradient datastore, substantially reducing storage costs while maintaining effective influence computations for data valuation.

\paragraph{Quantized Random Projection}  
Quantized random projection (QRP) is a powerful technique for dimensionality reduction and efficient similarity search. Early works~\cite{JLBinaryCoins,QuantizedJL} established its theoretical foundations, demonstrating that QRP can preserve pairwise distances while significantly reducing storage requirements. Subsequent studies have refined these guarantees, exploring various quantization schemes and showing that even with aggressive compression, high-dimensional vectors (such as gradients) retain sufficient geometric information for accurate similarity computations~\cite{VerySparseRP,QRPCosine,QRPGuarantee}. Despite its success in other domains, QRP's application to gradient-based data valuation in large language models remains largely unexplored, motivating our development of QLESS.

\section{Conclusion}
\label{sec:discussion}
We propose QLESS, a quantization-enhanced data selection algorithm that extends LESS with gradient compression techniques (\textsection\ref{sec:methodology}). QLESS maintains selection effectiveness while dramatically reducing storage requirements through absmax-based quantization of gradient features (\textsection\ref{sec:methodology}). Our experiments in \textsection\ref{sec:experiments} demonstrate that even extreme quantization (1-bit) can achieve competitive performance compared to half-precision baselines across multiple model architectures. 
Looking forward, QLESS could be extended to support dynamic quantization schemes that adapt to gradient distributions during training. The surprising effectiveness of 1-bit representations also suggests investigating theoretical connections between gradient quantization and influence estimation.

\subsection{Limitations}
\label{sec:limitations}

Despite QLESS's strong performance, several limitations warrant discussion. First, while our method substantially reduces storage requirements, the initial gradient computation still entails significant computational overhead. Although using QLoRA can help mitigate this cost, especially for large language models, further research is necessary to lower the overall gradient extraction burden for extensive datasets. Second, our sequential compression approach—applying random projection followed by quantization—may not optimally preserve influence relationships, as joint optimization of dimensionality reduction and precision reduction could potentially retain more information. Finally, the effectiveness of different quantization schemes (e.g., absmax vs.\ absmean) can vary by bit width, model architecture, and dataset characteristics; thus, practitioners must carefully select and empirically validate suitable strategies for each new scenario.

\section{Impact Statement}

This paper presents work whose goal is to advance the field of 
Machine Learning. There are many potential societal consequences 
of our work, none which we feel must be specifically highlighted here.

\bibliography{example_paper}

\begin{thebibliography}{41}
\providecommand{\natexlab}[1]{#1}
\providecommand{\url}[1]{\texttt{#1}}
\expandafter\ifx\csname urlstyle\endcsname\relax
  \providecommand{\doi}[1]{doi: #1}\else
  \providecommand{\doi}{doi: \begingroup \urlstyle{rm}\Url}\fi

\bibitem[Achlioptas(2003)]{JLBinaryCoins}
Achlioptas, D.
\newblock Database-friendly random projections: Johnson-lindenstrauss with binary coins.
\newblock \emph{Journal of Computer and System Sciences}, 66\penalty0 (4):\penalty0 671--687, 2003.
\newblock ISSN 0022-0000.
\newblock \doi{https://doi.org/10.1016/S0022-0000(03)00025-4}.
\newblock URL \url{https://www.sciencedirect.com/science/article/pii/S0022000003000254}.
\newblock Special Issue on PODS 2001.

\bibitem[Aji \& Heafield(2017)Aji and Heafield]{SparseGD1}
Aji, A.~F. and Heafield, K.
\newblock Sparse communication for distributed gradient descent.
\newblock In Palmer, M., Hwa, R., and Riedel, S. (eds.), \emph{Proceedings of the 2017 Conference on Empirical Methods in Natural Language Processing}, pp.\  440--445, Copenhagen, Denmark, September 2017. Association for Computational Linguistics.
\newblock \doi{10.18653/v1/D17-1045}.
\newblock URL \url{https://aclanthology.org/D17-1045/}.

\bibitem[Alistarh et~al.(2017)Alistarh, Grubic, Li, Tomioka, and Vojnovic]{QSGD}
Alistarh, D., Grubic, D., Li, J.~Z., Tomioka, R., and Vojnovic, M.
\newblock Qsgd: communication-efficient sgd via gradient quantization and encoding.
\newblock In \emph{Proceedings of the 31st International Conference on Neural Information Processing Systems}, NIPS'17, pp.\  1707–1718, Red Hook, NY, USA, 2017. Curran Associates Inc.
\newblock ISBN 9781510860964.

\bibitem[Bernstein et~al.(2018)Bernstein, Wang, Azizzadenesheli, and Anandkumar]{SignSGD}
Bernstein, J., Wang, Y.-X., Azizzadenesheli, K., and Anandkumar, A.
\newblock signsgd: Compressed optimisation for non-convex problems.
\newblock In \emph{International Conference on Machine Learning}, pp.\  560--569. PMLR, 2018.

\bibitem[Charikar(2002)]{Simhash}
Charikar, M.~S.
\newblock Similarity estimation techniques from rounding algorithms.
\newblock In \emph{Proceedings of the Thiry-Fourth Annual ACM Symposium on Theory of Computing}, STOC '02, pp.\  380–388, New York, NY, USA, 2002. Association for Computing Machinery.
\newblock ISBN 1581134959.
\newblock \doi{10.1145/509907.509965}.
\newblock URL \url{https://doi.org/10.1145/509907.509965}.

\bibitem[Choe et~al.(2024)Choe, Ahn, Bae, Zhao, Kang, Chung, Pratapa, Neiswanger, Strubell, Mitamura, Schneider, Hovy, Grosse, and Xing]{loGra}
Choe, S.~K., Ahn, H., Bae, J., Zhao, K., Kang, M., Chung, Y., Pratapa, A., Neiswanger, W., Strubell, E., Mitamura, T., Schneider, J., Hovy, E., Grosse, R., and Xing, E.
\newblock What is your data worth to gpt? llm-scale data valuation with influence functions, 2024.
\newblock URL \url{https://arxiv.org/abs/2405.13954}.

\bibitem[Clark et~al.(2020)Clark, Choi, Collins, Garrette, Kwiatkowski, Nikolaev, and Palomaki]{tydiqa}
Clark, J.~H., Choi, E., Collins, M., Garrette, D., Kwiatkowski, T., Nikolaev, V., and Palomaki, J.
\newblock {TyDi QA}: A benchmark for information-seeking question answering in typologically diverse languages.
\newblock \emph{Transactions of the Association for Computational Linguistics}, 2020.

\bibitem[Conover et~al.(2023)Conover, Hayes, Mathur, Xie, Wan, Shah, Ghodsi, Wendell, Zaharia, and Xin]{Dolly}
Conover, M., Hayes, M., Mathur, A., Xie, J., Wan, J., Shah, S., Ghodsi, A., Wendell, P., Zaharia, M., and Xin, R.
\newblock Free {Dolly}: Introducing the world's first truly open instruction-tuned {LLM}, 2023.

\bibitem[Dettmers(2016)]{Grad8bit}
Dettmers, T.
\newblock 8-bit approximations for parallelism in deep learning.
\newblock In \emph{ICLR (Poster)}, 2016.
\newblock URL \url{http://arxiv.org/abs/1511.04561}.

\bibitem[Dettmers et~al.(2022)Dettmers, Lewis, Belkada, and Zettlemoyer]{LLMint8}
Dettmers, T., Lewis, M., Belkada, Y., and Zettlemoyer, L.
\newblock Llm.int8(): 8-bit matrix multiplication for transformers at scale, 2022.
\newblock URL \url{https://arxiv.org/abs/2208.07339}.

\bibitem[Dettmers et~al.(2024)Dettmers, Pagnoni, Holtzman, and Zettlemoyer]{QLoRa}
Dettmers, T., Pagnoni, A., Holtzman, A., and Zettlemoyer, L.
\newblock Qlora: efficient finetuning of quantized llms.
\newblock In \emph{Proceedings of the 37th International Conference on Neural Information Processing Systems}, NIPS '23, Red Hook, NY, USA, 2024. Curran Associates Inc.

\bibitem[Dubey et~al.(2024)Dubey, Jauhri, Pandey, Kadian, Al-Dahle, Letman, Mathur, Schelten, Yang, Fan, et~al.]{Llama3}
Dubey, A., Jauhri, A., Pandey, A., Kadian, A., Al-Dahle, A., Letman, A., Mathur, A., Schelten, A., Yang, A., Fan, A., et~al.
\newblock The llama 3 herd of models.
\newblock \emph{arXiv preprint arXiv:2407.21783}, 2024.

\bibitem[Garima et~al.(2020)Garima, Liu, Kale, and Sundararajan]{tracin}
Garima, Liu, F., Kale, S., and Sundararajan, M.
\newblock Estimating training data influence by tracing gradient descent.
\newblock In \emph{Proceedings of the 34th International Conference on Neural Information Processing Systems}, NIPS '20, Red Hook, NY, USA, 2020. Curran Associates Inc.
\newblock ISBN 9781713829546.

\bibitem[Hendrycks et~al.(2020)Hendrycks, Burns, Basart, Zou, Mazeika, Song, and Steinhardt]{mmlu}
Hendrycks, D., Burns, C., Basart, S., Zou, A., Mazeika, M., Song, D., and Steinhardt, J.
\newblock Measuring massive multitask language understanding.
\newblock In \emph{International Conference on Learning Representations}, 2020.

\bibitem[Hu et~al.(2021)Hu, Shen, Wallis, Allen-Zhu, Li, Wang, Wang, and Chen]{LoRA}
Hu, E.~J., Shen, Y., Wallis, P., Allen-Zhu, Z., Li, Y., Wang, S., Wang, L., and Chen, W.
\newblock Lora: Low-rank adaptation of large language models.
\newblock \emph{arXiv preprint arXiv:2106.09685}, 2021.

\bibitem[Hyeon-Woo et~al.(2022)Hyeon-Woo, Ye-Bin, and Oh]{LoHa}
Hyeon-Woo, N., Ye-Bin, M., and Oh, T.-H.
\newblock Fedpara: Low-rank hadamard product for communication-efficient federated learning.
\newblock In \emph{International Conference on Learning Representations}, 2022.
\newblock URL \url{https://openreview.net/forum?id=d71n4ftoCBy}.

\bibitem[Jacques(2015)]{QuantizedJL}
Jacques, L.
\newblock A quantized johnson–lindenstrauss lemma: The finding of buffon’s needle.
\newblock \emph{IEEE Transactions on Information Theory}, 61\penalty0 (9):\penalty0 5012--5027, 2015.
\newblock \doi{10.1109/TIT.2015.2453355}.

\bibitem[Jacques \& Cambareri(2017)Jacques and Cambareri]{QRPGuarantee}
Jacques, L. and Cambareri, V.
\newblock Time for dithering: fast and quantized random embeddings via the restricted isometry property.
\newblock \emph{Information and Inference: A Journal of the IMA}, 6\penalty0 (4):\penalty0 441--476, 04 2017.
\newblock ISSN 2049-8764.
\newblock \doi{10.1093/imaiai/iax004}.
\newblock URL \url{https://doi.org/10.1093/imaiai/iax004}.

\bibitem[Jiang et~al.(2023)Jiang, Sablayrolles, Mensch, Bamford, Chaplot, Casas, Bressand, Lengyel, Lample, Saulnier, et~al.]{Mistral7B}
Jiang, A.~Q., Sablayrolles, A., Mensch, A., Bamford, C., Chaplot, D.~S., Casas, D. d.~l., Bressand, F., Lengyel, G., Lample, G., Saulnier, L., et~al.
\newblock Mistral 7b.
\newblock \emph{arXiv preprint arXiv:2310.06825}, 2023.

\bibitem[Johnson \& Lindenstrauss(1984)Johnson and Lindenstrauss]{JohnsonLindenstrauss}
Johnson, W.~B. and Lindenstrauss, J.
\newblock Extensions of lipschitz mappings into hilbert space.
\newblock \emph{Contemporary mathematics}, 26:\penalty0 189--206, 1984.

\bibitem[Koh \& Liang(2017)Koh and Liang]{influence}
Koh, P.~W. and Liang, P.
\newblock Understanding black-box predictions via influence functions.
\newblock In \emph{International conference on machine learning}, pp.\  1885--1894. PMLR, 2017.

\bibitem[K{\"o}pf et~al.(2023)K{\"o}pf, Kilcher, von R{\"u}tte, Anagnostidis, Tam, Stevens, Barhoum, Duc, Stanley, Nagyfi, et~al.]{OpenAssistant}
K{\"o}pf, A., Kilcher, Y., von R{\"u}tte, D., Anagnostidis, S., Tam, Z.-R., Stevens, K., Barhoum, A., Duc, N.~M., Stanley, O., Nagyfi, R., et~al.
\newblock {OpenAssistant} conversations--democratizing large language model alignment.
\newblock 2023.

\bibitem[Kwon et~al.(2024)Kwon, Wu, Wu, and Zou]{DataInf}
Kwon, Y., Wu, E., Wu, K., and Zou, J.
\newblock Datainf: Efficiently estimating data influence in lo{RA}-tuned {LLM}s and diffusion models.
\newblock In \emph{The Twelfth International Conference on Learning Representations}, 2024.
\newblock URL \url{https://openreview.net/forum?id=9m02ib92Wz}.

\bibitem[Li et~al.(2012)Li, Rane, and Boufounos]{QRPImage}
Li, M., Rane, S., and Boufounos, P.
\newblock Quantized embeddings of scale-invariant image features for mobile augmented reality.
\newblock In \emph{2012 IEEE 14th International Workshop on Multimedia Signal Processing (MMSP)}, pp.\  1--6, 2012.
\newblock \doi{10.1109/MMSP.2012.6343406}.

\bibitem[Li et~al.(2006)Li, Hastie, and Church]{VerySparseRP}
Li, P., Hastie, T.~J., and Church, K.~W.
\newblock Very sparse random projections.
\newblock In \emph{Proceedings of the 12th ACM SIGKDD International Conference on Knowledge Discovery and Data Mining}, KDD '06, pp.\  287–296, New York, NY, USA, 2006. Association for Computing Machinery.
\newblock ISBN 1595933395.
\newblock \doi{10.1145/1150402.1150436}.
\newblock URL \url{https://doi.org/10.1145/1150402.1150436}.

\bibitem[Li et~al.(2014)Li, Mitzenmacher, and Shrivastava]{RPCoding}
Li, P., Mitzenmacher, M., and Shrivastava, A.
\newblock Coding for random projections.
\newblock In Xing, E.~P. and Jebara, T. (eds.), \emph{Proceedings of the 31st International Conference on Machine Learning}, volume~32 of \emph{Proceedings of Machine Learning Research}, pp.\  676--684, Bejing, China, 22--24 Jun 2014. PMLR.
\newblock URL \url{https://proceedings.mlr.press/v32/lie14.html}.

\bibitem[Li et~al.(2016)Li, Mitzenmacher, and Slawski]{QRPCosine}
Li, P., Mitzenmacher, M., and Slawski, M.
\newblock Quantized random projections and non-linear estimation of cosine similarity.
\newblock In Lee, D., Sugiyama, M., Luxburg, U., Guyon, I., and Garnett, R. (eds.), \emph{Advances in Neural Information Processing Systems}, volume~29. Curran Associates, Inc., 2016.
\newblock URL \url{https://proceedings.neurips.cc/paper_files/paper/2016/file/186a157b2992e7daed3677ce8e9fe40f-Paper.pdf}.

\bibitem[Liu et~al.(2024)Liu, Qiu, Feng, Xiu, Xue, Yu, Feng, Liu, Heo, Peng, Wen, Black, Weller, and Sch{\"o}lkopf]{BOFT}
Liu, W., Qiu, Z., Feng, Y., Xiu, Y., Xue, Y., Yu, L., Feng, H., Liu, Z., Heo, J., Peng, S., Wen, Y., Black, M.~J., Weller, A., and Sch{\"o}lkopf, B.
\newblock Parameter-efficient orthogonal finetuning via butterfly factorization.
\newblock In \emph{The Twelfth International Conference on Learning Representations}, 2024.
\newblock URL \url{https://openreview.net/forum?id=7NzgkEdGyr}.

\bibitem[Longpre et~al.(2023)Longpre, Hou, Vu, Webson, Chung, Tay, Zhou, Le, Zoph, Wei, et~al.]{flanv2}
Longpre, S., Hou, L., Vu, T., Webson, A., Chung, H.~W., Tay, Y., Zhou, D., Le, Q.~V., Zoph, B., Wei, J., et~al.
\newblock The flan collection: Designing data and methods for effective instruction tuning.
\newblock \emph{arXiv preprint arXiv:2301.13688}, 2023.

\bibitem[Park et~al.(2023)Park, Georgiev, Ilyas, Leclerc, and Madry]{trak}
Park, S.~M., Georgiev, K., Ilyas, A., Leclerc, G., and Madry, A.
\newblock Trak: attributing model behavior at scale.
\newblock In \emph{Proceedings of the 40th International Conference on Machine Learning}, ICML'23. JMLR.org, 2023.

\bibitem[Qiu et~al.(2023)Qiu, Liu, Feng, Xue, Feng, Liu, Zhang, Weller, and Sch{\"o}lkopf]{OFT}
Qiu, Z., Liu, W., Feng, H., Xue, Y., Feng, Y., Liu, Z., Zhang, D., Weller, A., and Sch{\"o}lkopf, B.
\newblock Controlling text-to-image diffusion by orthogonal finetuning.
\newblock In \emph{Thirty-seventh Conference on Neural Information Processing Systems}, 2023.
\newblock URL \url{https://openreview.net/forum?id=K30wTdIIYc}.

\bibitem[Qwen et~al.(2025)Qwen, :, Yang, Yang, Zhang, Hui, Zheng, Yu, Li, Liu, Huang, Wei, Lin, Yang, Tu, Zhang, Yang, Yang, Zhou, Lin, Dang, Lu, Bao, Yang, Yu, Li, Xue, Zhang, Zhu, Men, Lin, Li, Tang, Xia, Ren, Ren, Fan, Su, Zhang, Wan, Liu, Cui, Zhang, and Qiu]{Qwen25}
Qwen, :, Yang, A., Yang, B., Zhang, B., Hui, B., Zheng, B., Yu, B., Li, C., Liu, D., Huang, F., Wei, H., Lin, H., Yang, J., Tu, J., Zhang, J., Yang, J., Yang, J., Zhou, J., Lin, J., Dang, K., Lu, K., Bao, K., Yang, K., Yu, L., Li, M., Xue, M., Zhang, P., Zhu, Q., Men, R., Lin, R., Li, T., Tang, T., Xia, T., Ren, X., Ren, X., Fan, Y., Su, Y., Zhang, Y., Wan, Y., Liu, Y., Cui, Z., Zhang, Z., and Qiu, Z.
\newblock Qwen2.5 technical report, 2025.
\newblock URL \url{https://arxiv.org/abs/2412.15115}.

\bibitem[Stich et~al.(2018)Stich, Cordonnier, and Jaggi]{SparseGD3}
Stich, S.~U., Cordonnier, J.-B., and Jaggi, M.
\newblock Sparsified sgd with memory.
\newblock In \emph{Proceedings of the 32nd International Conference on Neural Information Processing Systems}, NIPS'18, pp.\  4452–4463, Red Hook, NY, USA, 2018. Curran Associates Inc.

\bibitem[Suzgun et~al.(2023)Suzgun, Scales, Sch{\"a}rli, Gehrmann, Tay, Chung, Chowdhery, Le, Chi, Zhou, et~al.]{bbh}
Suzgun, M., Scales, N., Sch{\"a}rli, N., Gehrmann, S., Tay, Y., Chung, H.~W., Chowdhery, A., Le, Q., Chi, E., Zhou, D., et~al.
\newblock Challenging big-bench tasks and whether chain-of-thought can solve them.
\newblock In \emph{Findings of the Association for Computational Linguistics: ACL 2023}, pp.\  13003--13051, 2023.

\bibitem[Touvron et~al.(2023)Touvron, Martin, Stone, Albert, Almahairi, Babaei, Bashlykov, Batra, Bhargava, Bhosale, et~al.]{Llama2}
Touvron, H., Martin, L., Stone, K., Albert, P., Almahairi, A., Babaei, Y., Bashlykov, N., Batra, S., Bhargava, P., Bhosale, S., et~al.
\newblock Llama 2: Open foundation and fine-tuned chat models.
\newblock \emph{arXiv preprint arXiv:2307.09288}, 2023.

\bibitem[Wangni et~al.(2018)Wangni, Wang, Liu, and Zhang]{SparseGD2}
Wangni, J., Wang, J., Liu, J., and Zhang, T.
\newblock Gradient sparsification for communication-efficient distributed optimization.
\newblock In Bengio, S., Wallach, H., Larochelle, H., Grauman, K., Cesa-Bianchi, N., and Garnett, R. (eds.), \emph{Advances in Neural Information Processing Systems}, volume~31. Curran Associates, Inc., 2018.
\newblock URL \url{https://proceedings.neurips.cc/paper_files/paper/2018/file/3328bdf9a4b9504b9398284244fe97c2-Paper.pdf}.

\bibitem[Wei et~al.(2022)Wei, Wang, Schuurmans, Bosma, Xia, Chi, Le, Zhou, et~al.]{COT}
Wei, J., Wang, X., Schuurmans, D., Bosma, M., Xia, F., Chi, E., Le, Q.~V., Zhou, D., et~al.
\newblock Chain-of-thought prompting elicits reasoning in large language models.
\newblock \emph{Advances in Neural Information Processing Systems}, 35:\penalty0 24824--24837, 2022.

\bibitem[Wen et~al.(2017)Wen, Xu, Yan, Wu, Wang, Chen, and Li]{TernGrad}
Wen, W., Xu, C., Yan, F., Wu, C., Wang, Y., Chen, Y., and Li, H.
\newblock Terngrad: ternary gradients to reduce communication in distributed deep learning.
\newblock In \emph{Proceedings of the 31st International Conference on Neural Information Processing Systems}, NIPS'17, pp.\  1508–1518, Red Hook, NY, USA, 2017. Curran Associates Inc.
\newblock ISBN 9781510860964.

\bibitem[Xia et~al.(2025)Xia, Malladi, Gururangan, Arora, and Chen]{less}
Xia, M., Malladi, S., Gururangan, S., Arora, S., and Chen, D.
\newblock Less: selecting influential data for targeted instruction tuning.
\newblock In \emph{Proceedings of the 41st International Conference on Machine Learning}, ICML'24. JMLR.org, 2025.

\bibitem[YEH et~al.(2024)YEH, Hsieh, Gao, Yang, Oh, and Gong]{LoKr}
YEH, S.-Y., Hsieh, Y.-G., Gao, Z., Yang, B. B.~W., Oh, G., and Gong, Y.
\newblock Navigating text-to-image customization: From ly{CORIS} fine-tuning to model evaluation.
\newblock In \emph{The Twelfth International Conference on Learning Representations}, 2024.
\newblock URL \url{https://openreview.net/forum?id=wfzXa8e783}.

\bibitem[Zhang et~al.(2023)Zhang, Chen, Bukharin, He, Cheng, Chen, and Zhao]{AdaLoRA}
Zhang, Q., Chen, M., Bukharin, A., He, P., Cheng, Y., Chen, W., and Zhao, T.
\newblock Adaptive budget allocation for parameter-efficient fine-tuning.
\newblock In \emph{The Eleventh International Conference on Learning Representations}, 2023.
\newblock URL \url{https://openreview.net/forum?id=lq62uWRJjiY}.

\end{thebibliography}
\bibliographystyle{icml2025}

%%%%%%%%%%%%%%%%%%%%%%%%%%%%%%%%%%%%%%%%%%%%%%%%%%%%%%%%%%%%%%%%%%%%%%%%%%%%%%%
%%%%%%%%%%%%%%%%%%%%%%%%%%%%%%%%%%%%%%%%%%%%%%%%%%%%%%%%%%%%%%%%%%%%%%%%%%%%%%%
% APPENDIX
%%%%%%%%%%%%%%%%%%%%%%%%%%%%%%%%%%%%%%%%%%%%%%%%%%%%%%%%%%%%%%%%%%%%%%%%%%%%%%%
%%%%%%%%%%%%%%%%%%%%%%%%%%%%%%%%%%%%%%%%%%%%%%%%%%%%%%%%%%%%%%%%%%%%%%%%%%%%%%%
\newpage
\appendix
\onecolumn
\section{Experimental Details}
\label{app:experimental_details}

All experiments were conducted using the parameter-efficient fine-tuning method LoRA~\cite{LoRA}. We employed a learning rate scheduler with linear warm-up and cosine decay, reaching a peak learning rate of $2 \times 10^{-5}$. Training was carried out for 4 epochs across all selected datasets with a batch size of 128. For the LoRA module, we specified a rank of 64, an $\alpha$ value of 256, a dropout rate of 0.1, and learned LoRA matrices for query, key, value, and output projection layers. 

We performed three trials using distinct random seeds. For random selection approaches, this involved selecting three different random subsets from the training dataset. In our approach (QLESS), this meant conducting warmup training with various subsets of the training data and subsequently selecting different subsets for each trial from each warmup-trained model. We maintained consistent optimization seeds across all experiments to ensure reproducibility.

We followed an evaluation procedure similar to \cite{less}, focusing on three target tasks. 
For MMLU, we reported 5-shot accuracy on the test set, averaging results across 57 subtasks. 
For TyDiQA, we measured 1-shot macro-averaged F1 over all 11 languages using the gold-passage setup. 
For BBH, we computed the 3-shot exact match score across all tasks, providing chain-of-thought reasoning in each in-context learning example. 

All experiments were conducted on 4 NVIDIA A100 GPUs with 40GB memory. We utilized BFloat16 precision and employed Fully Sharded Data Parallel (FSDP) training with auto-wrapping to optimize memory and computational efficiency. The per-device batch size was set to 1, with gradient accumulation steps adjusted to achieve an effective batch size of 128.

\section{Additional Results}
\label{app:additional-results}

To further validate the effectiveness and generalizability of QLESS, we conducted experiments on two additional large language models: Llama 3.2 3B and Mistral 7B. Table~\ref{table:additional-model-table} presents the performance of these models, as well as the original Llama 2 7B, across different data selection methods and gradient storage configurations. Table~\ref{tab:llama-2-7b-quant} presents the performance of Llama 2 7B  under different model and gradient quantization settings.

\begin{table*}[ht!]
\caption{Performance of Llama 2 7B, Mistral 7B, and Llama 3.2 3B models with different data selection methods and gradient storage space. Metrics include TyDiQA, MMLU, BBH, and their average in percent (\%). \textbf{Bold} and \underline{underline} denote the best and second best performing setup across selection methods of the same model.}
\label{table:additional-model-table}
\centering
\small
\vspace{3mm}
\begin{tabular}{lccccccc}
\toprule
\textbf{Model} & \textbf{Data Selection} & \textbf{TyDiQA} & \textbf{MMLU} & \textbf{BBH} & \textbf{Avg} \\
\midrule
    \multirow{7}{*}{Llama 2 7B} & random 100\% & 49.89 \std{0.0} & \textbf{47.30} \std{0.0} & \textbf{40.93} \std{0.0} & \textbf{46.04} \std{0.0} \\ 
    ~ & random 5\% & 47.94 \std{0.2} & 45.73 \std{0.6} & \underline{39.57} \std{0.7} & 44.41 \std{0.4} \\ 
    ~ & LESS 16-bit & \underline{51.21} \std{4.0} & 45.80 \std{1.6} & 38.43 \std{2.0} & 45.15 \std{2.4} \\ 
    ~ & QLESS 8-bit & \textbf{51.40} \std{5.6} & 45.73 \std{1.4} & 38.58 \std{1.7} & \underline{45.24} \std{2.9} \\ 
    ~ & QLESS 4-bit & 50.55 \std{4.6} & 45.80 \std{1.5} & 38.73 \std{1.4} & 45.03 \std{2.4} \\ 
    ~ & QLESS 2-bit & 49.95 \std{3.9} & 45.53 \std{1.1} & 38.27 \std{2.0} & 44.59 \std{2.3} \\ 
    ~ & QLESS 1-bit & 50.61 \std{5.5} & \underline{45.97} \std{1.8} & 38.49 \std{1.7} & 45.02 \std{3.0} \\ 
    \midrule
    \multirow{7}{*}{Mistral 7B} & random 100\% & \textbf{62.78} \std{0.0} & 61.10 \std{0.0} & 56.76 \std{0.0} & \textbf{60.21} \std{0.0} \\ 
    ~ & random 5\% & \underline{60.31} \std{2.0} & 60.97 \std{0.1} & 56.94 \std{1.2} & 59.41 \std{1.0} \\ 
    ~ & LESS 16-bit & 58.00 \std{6.3} & 60.87 \std{0.6} & \textbf{60.22} \std{6.7} & \underline{59.70} \std{2.5} \\ 
    ~ & QLESS 8-bit & 57.83 \std{4.5} & \underline{61.17} \std{1.4} & 56.36 \std{0.3} & 58.45 \std{1.9} \\ 
    ~ & QLESS 4-bit & 57.10 \std{5.5} & 60.50 \std{0.4} & 55.68 \std{1.0} & 57.76 \std{1.6} \\ 
    ~ & QLESS 2-bit & 57.51 \std{3.3} & 60.73 \std{0.7} & \underline{57.31} \std{0.5} & 58.52 \std{1.2} \\ 
    ~ & QLESS 1-bit & 58.60 \std{5.7} & \textbf{61.70} \std{0.3} & 55.46 \std{0.3} & 58.59 \std{1.9} \\ 
    \midrule
    \multirow{7}{*}{Llama 3.2 3B} & random 100\% & 59.76 \std{0.0} & 54.80 \std{0.0} & 38.32 \std{0.0} & 50.96 \std{0.0} \\ 
    ~ & random 5\% & 60.86 \std{1.7} & \textbf{55.83} \std{0.1} & \textbf{48.55} \std{0.9} & 55.08 \std{0.6} \\ 
    ~ & LESS 16-bit & \textbf{63.27} \std{4.3} & \underline{55.70} \std{0.2} & 46.27 \std{1.4} & 55.08 \std{1.9} \\ 
    ~ & QLESS 8-bit & \underline{63.12} \std{2.9} & 55.63 \std{0.2} & 47.16 \std{1.5} & \textbf{55.31} \std{1.4} \\ 
    ~ & QLESS 4-bit & 62.05 \std{4.4} & 55.67 \std{0.1} & 46.79 \std{0.3} & 54.83 \std{1.6} \\ 
    ~ & QLESS 2-bit & 61.30 \std{0.6} & 55.80 \std{0.2} & 46.48 \std{2.0} & 54.53 \std{0.9} \\ 
    ~ & QLESS 1-bit & 62.59 \std{4.2} & \underline{55.70} \std{0.1} & \underline{47.28} \std{1.6} & \underline{55.19} \std{1.9} \\ 
\bottomrule
\end{tabular}
\end{table*}

\begin{table*}[ht!]
\caption{Performance of Llama 2 7B under different model and gradient quantization settings across TyDiQA, MMLU, and BBH. The final column presents the average performance across these tasks. \textbf{Bold} and \underline{underline} denote the best and second best performing setup across model quantization and gradient quantization.}
\label{tab:llama-2-7b-quant}
\centering
\small
\vspace{3mm}
\begin{tabular}{@{ }c@{ }cHcccc@{ }} 
\toprule
\textbf{Model Q} & \textbf{Grad Q} & \textbf{Memory} & \textbf{TyDiQA} & \textbf{MMLU} & \textbf{BBH} & \textbf{Avg} \\ 
\textbf{(Mem.)} \\ 
\midrule
– & rand 100\% & – & 49.89 & 47.30 & 40.93 & 46.04  \\
– & rand 5\% & – & 48.13 & 45.00 & 38.70 & 43.94 \\ 
\cmidrule(lr){1-7}
\multirow{5}{*}{\shortstack{16-bit \\ (32.35 GB)}} & 16-bit & \multirow{5}{*}{32.35 GB} & 54.97 & 47.70 & 40.74 & 47.80 \\
 & 8-bit &  & \underline{57.81} & 47.40 & 40.56 & \underline{48.59} \\
 & 4-bit &  & 55.75 & 47.50 & 40.28 & 47.84 \\
 & 2-bit &  & 54.48 & 46.80 & 40.46 & 47.25 \\
 & 1-bit &  & 56.93 & \textbf{48.00} & 40.46 & 48.46 \\ 
\cmidrule(lr){1-7}
\multirow{5}{*}{\shortstack{8-bit \\ (31.75 GB)}} & 16-bit & \multirow{5}{*}{31.75 GB} & 55.24 & \underline{47.80} & \underline{41.67} & 48.24 \\
 & 8-bit &  & 53.34 & 47.70 & 40.83 & 47.29 \\
 & 4-bit &  & 57.07 & 47.50 & 40.56 & 48.38 \\
 & 2-bit &  & 53.70 & 47.20 & 39.72 & 46.88 \\
 & 1-bit &  & 55.04 & 47.60 & 41.20 & 47.95 \\ 
\cmidrule(lr){1-7}
\multirow{5}{*}{\shortstack{4-bit \\ (22.51 GB)}} & 16-bit & \multirow{5}{*}{22.51 GB} & 55.47 & 46.70 & 40.83 & 47.67 \\
 & 8-bit &  & 55.09 & 46.30 & \textbf{42.50} & 47.96 \\
 & 4-bit &  & 55.88 & 46.30 & 40.19 & 47.46 \\
 & 2-bit &  & 53.93 & 44.90 & 37.50 & 45.44 \\
 & 1-bit &  & \textbf{58.14} & 46.70 & 41.30 & \textbf{48.71} \\
\bottomrule
\end{tabular}
\end{table*}

\section{Qualitative Analysis }
\label{app:qualitative_analysis}
\begin{figure*}[h]
    \centering
    \subfigure[TyDiQA]{
        \label{fig:tydiqa_dist}
        \includegraphics[width=0.3\columnwidth]{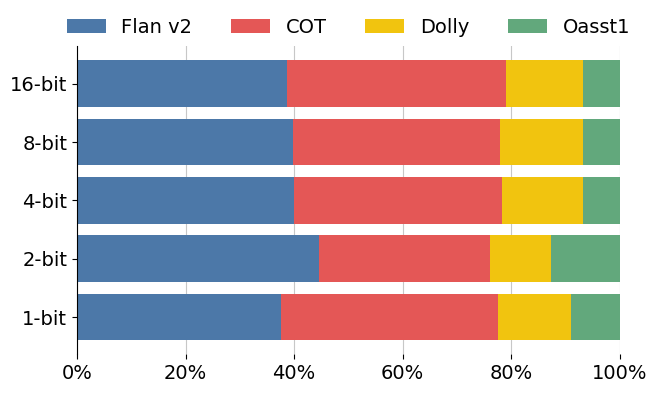}
    }
    \subfigure[MMLU]{
        \label{fig:mmlu_dist}
        \includegraphics[width=0.3\columnwidth]{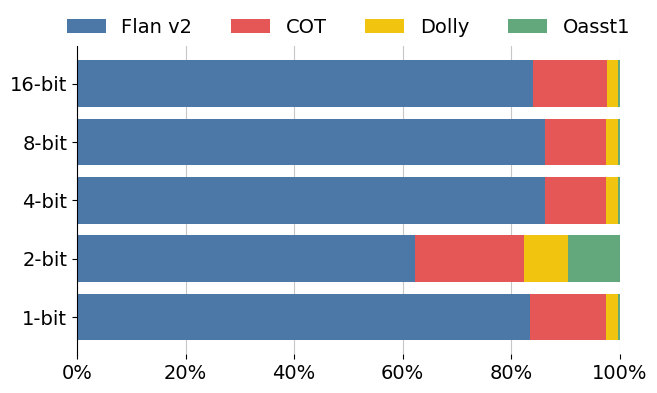}
    }
    \subfigure[BBH]{
        \label{fig:bbh_dist}
        \includegraphics[width=0.3\columnwidth]{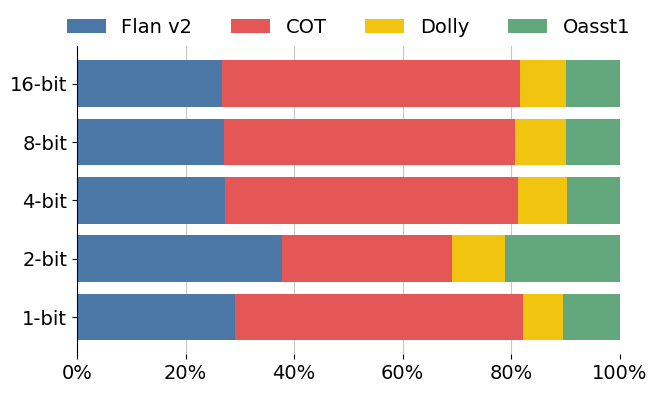}
    }
    \caption{Subset distribution of the top $5\%$ selected examples for different quantization levels.}
    \label{fig:task_dist}   
\end{figure*}

In addition to the quantitative metrics reported above, we examine which specific examples are chosen by QLESS under various quantization levels. Figure~\ref{fig:task_dist} shows the distribution of the top 5\% selected examples (by influence score) across four instructional datasets (Flan v2, CoT, Dolly, Oasst1) for three validation benchmarks (TyDiQA, MMLU, and BBH). Meanwhile, Tables~\ref{tab:tydiqa_example}, \ref{tab:mmlu_example}, and~\ref{tab:bbh_example} present concrete examples selected for each benchmark, illustrating how the chosen samples differ across bit-precision settings.

\paragraph{Distribution of Data Sources.}
From Figure~\ref{fig:task_dist}, we observe that the overall composition of the top 5\% selected training data remains relatively stable across 16-bit, 8-bit, and 4-bit, and even 1-bit quantizations, showing robustness over extreme quantization. The proportion of each dataset (Flan v2, CoT, Dolly, Oasst1) shift more noticeably at 2-bit, which may be caused by the sparsity effect of quantization (Figure~\ref{fig:absmean-v-absmax}). For instance, in BBH (Figure~\ref{fig:task_dist}\,(C)), 2-bit selection draws a larger fraction of examples from  Oasst1.

\paragraph{Granular Example Differences.}
For TyDiQA (Table \ref{tab:tydiqa_example}), we observe that the 16-bit, 8-bit, and 4-bit variants of QLESS select a highly relevant example that demonstrates the model's ability to extract factual information from a given passage to answer a question. The 2-bit and 1-bit variants, on the other hand, select examples that test the model's capacity to combine facts from different sources to answer more complex questions. While these examples are less directly aligned with the original validation example, they still assess important language understanding capabilities.
In the case of MMLU (Table \ref{tab:mmlu_example}), the 16-bit, 8-bit, 4-bit, and 1-bit QLESS variants all select a highly relevant example that involves solving a mathematical problem and selecting the correct multiple-choice answer. This aligns well with the problem-solving nature of the original validation example. Interestingly, the 2-bit variant selects a more creative example that tests the model's ability to generate a descriptive prompt for an image generation task. While less directly relevant, this example still challenges the model's language understanding and generation abilities.
Finally, for BBH (Table \ref{tab:bbh_example}), all QLESS variants select examples that involve reasoning through a multi-step problem to arrive at a numerical answer. The 16-bit, 8-bit, 4-bit, and 1-bit variants choose a highly relevant example that closely mirrors the step-by-step arithmetic reasoning required in the original validation example. The 2-bit variant, while still testing multi-step reasoning, selects an example that is more focused on logical deduction based on scientific facts.

\paragraph{Interpretation.}
These observations indicate that QLESS largely preserves consistent types of top-ranked examples at 16-bit, 8-bit, 4-bit, and even 1-bit. Only the 2-bit quantization stands out as notably different, likely because the jump in sparsity degrades data valuation precision enough to emphasize distinct prompts. Although our quantitative results in earlier sections show that 2-bit can still perform reasonably well, the shift in selected data highlights the delicate balance between extreme compression and influence fidelity. These qualitative findings reinforce the notion that QLESS is robust to quantization—even down to 1-bit—while hinting that 2-bit may require closer calibration to avoid undesired selection biases.

\begin{table*}
        \caption{Selected sample for TyDiQA validation data}
    
    \centering
    \resizebox{\textwidth}{!}{
    \begin{tabular}{@{}p{16cm}@{}}
        \toprule

        \multicolumn{1}{c}{\textbf{TyDiQA Validation Example}} \\ \midrule
        {\textless{}\textbar{}user\textbar{}\textgreater{} \newline
        Answer the following question based on the information in the given passage.Passage: Home Box Office (HBO) is an American premium cable and satellite television network that is owned by the namesake unit Home Box Office, Inc., a division of AT\&T's WarnerMedia. The program which featured on the network consists primarily of theatrically released motion pictures and original television shows, along with made-for-cable movies, documentaries and occasional comedy and concert specials.\newline
        Question: Who owns HBO? \newline \textless{}\textbar{}assistant\textbar{}\textgreater{} \newline
        Answer: Home Box Office}  \\ 
        
        \midrule\multicolumn{1}{c}{\textbf{Selected by 16-bit, 8-bit, 4-bit}} \\ \midrule
         {\textless{}\textbar{}user\textbar{}\textgreater{} \newline
         Where is Landscape Arch? Landscape Arch is a natural rock arch in Arches National Park in the U.S. state of Utah. With a length of 290.1 feet (88.4 m), it is the longest natural arch in the park and the United States, and the fifth-longest in the world. Landscape Arch was named by Frank Beckwith, who explored the area in the winter of 1933–34 as the leader of a scientific expedition. \newline
         \textless{}\textbar{}assistant\textbar{}\textgreater{} \newline
         Answer: Landscape Arch is a natural rock arch in Arches National Park in the U.S. state of Utah.} \\
         
         \midrule\multicolumn{1}{c}{\textbf{Selected by 2-bit}} \\ \midrule
         {\textless{}\textbar{}user\textbar{}\textgreater{} \newline
         Combine facts and answer this: What is the name of the actress who starred in Dreams of Gold: The Mel Fisher Story and *M*A*S*H*? Ans: Loretta Swit Combine facts and answer this: What actress starred in an American-Canadian drama television series between December 3, 2000 and August 7, 2005? Ans: best known for her roles as Melanie Marcus on Showtime's critically acclaimed series "Queer as Folk" Combine facts and answer this: Which magazine is distributed in more states, Parents or The Watchtower Ans: Parents Combine facts and answer this: Which is a constituency in Lincolnshire that was served by Sir Richard Winfrey as a member of Parliament from 1923-1924? \newline \textless{}\textbar{}assistant\textbar{}\textgreater{} \newline
         content"Ans: Gainsborough} \\
         
         \midrule\multicolumn{1}{c}{\textbf{Selected by 1-bit}} \\ \midrule
         {\textless{}\textbar{}user\textbar{}\textgreater{} \newline
         You need to complete the following task: Given these paragraphs about battery storage power stations, for how long a battery storage power station can output? A battery storage power station is a type of energy storage power station that uses a group of batteries to store electrical energy. Battery storage is the fastest responding dispatchable source of power on electric grids, and it is used to stabilise those grids, as battery storage can transition from standby to full power in under a second to deal with grid contingencies. At full rated power, battery storage power stations are generally designed to output for up to a few hours. Battery storage can be used for short-term peak power and ancillary services, such as providing operating reserve and frequency control to minimize the chance of power outages. They are often installed at, or close to, other active or disused power stations and may share the same grid connection to reduce costs. Since battery storage plants require no deliveries of fuel, are compact compared to generating stations and have no chimneys or large cooling systems, they can be rapidly installed and placed if necessary within urban areas, close to customer load. \newline
         \textless{}\textbar{}assistant\textbar{}\textgreater{} \newline
         Answer: At full rated power, battery storage power stations are generally designed to output for up to a few hours.}

           \\ \bottomrule
    \end{tabular}}
    \label{tab:tydiqa_example}
\end{table*}

\begin{table*}
        \caption{Selected sample for MMLU validation data}
    
    \centering
    \resizebox{\textwidth}{!}{
    \begin{tabular}{@{}p{16cm}@{}}
        \toprule

        \multicolumn{1}{c}{\textbf{MMLU Validation Example}} \\ \midrule
        {\textless{}\textbar{}user\textbar{}\textgreater{} \newline
        The following are multiple choice questions (with answers) about  high school mathematics.

        Joe was in charge of lights for a dance. The red light blinks every two seconds, the yellow light every three seconds, and the blue light every five seconds. If we include the very beginning and very end of the dance, how many times during a seven minute dance will all the lights come on at the same time? (Assume that all three lights blink simultaneously at the very beginning of the dance.) \newline
        A. 3 \newline
        B. 15 \newline
        C. 6 \newline
        D. 5 \newline
        Answer: \newline \textless{}\textbar{}assistant\textbar{}\textgreater{} \newline
        The answer is: B}  \\ 
        
        \midrule\multicolumn{1}{c}{\textbf{Selected by 16-bit, 8-bit, 4-bit, 1-bit}} \\ \midrule
         {\textless{}\textbar{}user\textbar{}\textgreater{} \newline
         In this task, you need to answer the given multiple-choice question on the gain. Gain is the value by which to multiply the input. Classify your answers into 'a', 'b', 'c', 'd', and 'e'. 
         Input: Consider Input: Problem: jaclyn buys \$ 10 000 worth of debentures in a company . she earns 9.5 \% p . a . simple interest , paid to her quarterly ( that is , every 3 months ) . if the agreed period of the debenture was 18 months : calculate the amount of interest jaclyn will earn for each quarter Options: \newline
         a ) 237.5 , \newline
         b ) 234 , \newline
         c ) 289.5 , \newline
         d ) 345 , \newline
         e ) none of these \newline
         Output: \newline
         \textless{}\textbar{}assistant\textbar{}\textgreater{} \newline
         Output: d} \\
         
         \midrule\multicolumn{1}{c}{\textbf{Selected by 2-bit}} \\ \midrule
         {\textless{}\textbar{}user\textbar{}\textgreater{} \newline
         Give me a text2img prompt to generate a Ghibli wallpaper of a cozy cabin in a forest \newline \textless{}\textbar{}assistant\textbar{}\textgreater{} \newline
         Studio Ghibli, Cozy Cabin, In A Forest, Bright Colors, Anime Illustration, Cinematic, Professional Lighting, Very Detailed Illustration, 8k} \\
         
           \\ \bottomrule
    \end{tabular}}
    \label{tab:mmlu_example}
\end{table*}

\begin{table*}
        \caption{Selected sample for BBH validation data}
    
    \centering
    \resizebox{\textwidth}{!}{
    \begin{tabular}{@{}p{16cm}@{}}
        \toprule

        \multicolumn{1}{c}{\textbf{BBH Validation Example}} \\ \midrule
        {\textless{}\textbar{}user\textbar{}\textgreater{} \newline
        Solve multi-step arithmetic problems. Q: ((-5 + 9 * -4 - 0) * (4 + -7 + 0 * -5)) = \newline \textless{}\textbar{}assistant\textbar{}\textgreater{} \newline
        A: Let's think step by step.
Let’s recall that the order of operations in mathematics is as follows: \newline
(1) Parentheses, \newline
(2) exponents,  \newline
(3) multiplication and division (from left to right), \newline
(4) addition and multiplication (from left to right).  \newline
So, remember to always compute the expressions inside parentheses or brackets first. \newline
This equation can be written as "A * B", where A = (-5 + 9 * -4 - 0) and B = (4 + -7 + 0 * -5). \newline
Let's calculate A = (-5 + 9 * -4 - 0) = (-5 + (9 * -4) - 0) = (-5 + (-36) - 0) = (-5 + -36 - 0)  = -5 - 36 = -41. \newline
Let's calculate B = (4 + -7 + 0 * -5) = (4 + -7 + (0 * -5)) = (4 + -7 + 0) = (4 + -7) = (4 - 7) = -3. \newline
Then, the final equation is A * B = -41 * -3 = (-61) * (-3) = 123. So the answer is 123.

}  \\ 
        
        \midrule\multicolumn{1}{c}{\textbf{Selected by 16-bit, 8-bit, 4-bit, 1-bit}} \\ \midrule
     {\textless{}\textbar{}user\textbar{}\textgreater{} \newline
I want to ask you a question. Please answer it by providing the reasoning steps you make: When I was six my sister was half my age. Now I'm 70, how old is my sister?
\newline \textless{}\textbar{}assistant\textbar{}\textgreater{} \newline
When you were six, your sister was half your age, so she was three years old. now you are 70, so to find out how old your sister is, we find how many years have passed since then: 70-6=64 Because she was 3, we can add those 64 years to her age to get her age now: 3+64 = 67 your sister is 67 years old now. \newline
\textless{}\textbar{}user\textbar{}\textgreater{} \newline
how confident are you with your answer?
\newline \textless{}\textbar{}assistant\textbar{}\textgreater{} \newline
I am confident in my answer. This is because the difference between your age and your sister's age never changes. Therefore, if she was three years younger than you (half your age) when you were 6 years old, she will still be three years younger than you (70 - 3 = 67) today.
     
}  
\\
         
 \midrule\multicolumn{1}{c}{\textbf{Selected by 2-bit}} \\ \midrule
 {\textless{}\textbar{}user\textbar{}\textgreater{} \newline
Ques and Ans: What can use energy in order to make food for itself and others? \newline
Options: - (A) organisms - (B) mollusks - (C) trees - (D) microbes - (E) seeds - (F) chlorophyll - (G) plants - (H) animals 
\newline (G) 
\newline Logic for the answer:
\newline \textless{}\textbar{}assistant\textbar{}\textgreater{} \newline
Producers use energy and inorganic molecules to make food. If the plant is green, it is a producer. Plants use energy to make food.

} \\
\\ \bottomrule

    \end{tabular}}
    \label{tab:bbh_example}
\end{table*}

%%%%%%%%%%%%%%%%%%%%%%%%%%%%%%%%%%%%%%%%%%%%%%%%%%%%%%%%%%%%%%%%%%%%%%%%%%%%%%%
%%%%%%%%%%%%%%%%%%%%%%%%%%%%%%%%%%%%%%%%%%%%%%%%%%%%%%%%%%%%%%%%%%%%%%%%%%%%%%%

\end{document}